\RequirePackage{fix-cm}
\documentclass[twocolumn]{svjour3} 
\smartqed 
\usepackage{graphicx}
\usepackage{algorithm}
\usepackage{algpseudocode}
\usepackage{subcaption} 
\usepackage{amsmath}
\usepackage{tabularx}

\usepackage{nccmath}
\usepackage{xcolor}
\usepackage{lipsum}
\usepackage{url}
\usepackage{eso-pic}

\AddToShipoutPictureBG{
\AtPageUpperLeft{%
    \raisebox{-3\baselineskip}{\makebox[\paperwidth]{\begin{minipage}{21cm}\centering
      \textcolor{red}{This is a pre-print of an article published in Intelligent Service Robotics.\\
      The final uthenticated version is available online at:\url{ https://doi.org/10.1007/s11370-019-00308-4}}
    \end{minipage}}}%
  }

}
\AddToShipoutPictureBG{
\AtPageLowerLeft{%
    \raisebox{3\baselineskip}{\makebox[\paperwidth]{\begin{minipage}{21cm}\centering
      \textcolor{red}{This is a pre-print of an article published in Intelligent Service Robotics.\\
      The final uthenticated version is available online at:\url{ https://doi.org/10.1007/s11370-019-00308-4}}
    \end{minipage}}}%
  }

}

\begin{document}

\title{Qualitative vision-based navigation based on sloped funnel lane concept\thanks{\textcolor{red}{This is a pre-print of an article published in Intelligent Service Robotics. The final uthenticated version is available online at:https://doi.org/10.1007/s11370-019-00308-4}}
}


\author{Mohamad Mahdi Kassir \and
Maziar Palhang \and Mohammad Reza Ahmadzadeh 
}


\institute{Mohamad Mahdi Kassir \at
Department of Electrical and Computer Engineering, Isfahan University of Technology, Isfahan, Iran \\
Tel.: +989126513834\\
Fax: +983133912451\\
\email{m.kaseer@ec.iut.ac.ir} 
\and
Maziar Palhang \at
Department of Electrical and Computer Engineering, Isfahan University of Technology, Isfahan, Iran \\
Tel.: +983133915426\\
Fax: +983133912451\\
\email{palhang@cc.iut.ac.ir}
\and
Mohammad Reza Ahmadzadeh \at
Department of Electrical and Computer Engineering, Isfahan University of Technology, Isfahan, Iran \\
Tel.: +983133915370\\
Fax: +983133912451\\
\email{ahmadzadeh@cc.iut.ac.ir}
}

\date{Received: date / Accepted: date}

\maketitle

\begin{abstract}
Funnel lane concept is a qualitative visual navigation method which helps robots to autonomously navigate by using a recorded video. A visual path is extracted from the video by extracting some keyframes from the video. The robot uses this visual path for its navigation. Funnel lane unlike some other methods does not make use of traditional calculations of Jacobians, homographies, fundamental matrices, or the focus of expansion, and does not require any camera calibration. However, funnel lane has some shortcomings. One problem is that funnel lane gives no information about the radius of rotation, so in turnings, the robot turns by a constant radius of rotation along the path. This reduces the maneuverability and limits the robot from dealing with all turnings conditions. In addition, this problem makes the robot faces a serious problem in correcting its path when it deviates from the desired path. Another flaw is that in some situations the robot faces an ambiguity to understand whether a translation or a rotation should be followed in the visual path which leads the robot to deviate and to fail in following the desired path. This paper introduces the sloped funnel lane technique which does not have these shortcomings. The roll and pitch angles are added to the funnel lane, which help the robot to set its radius of rotation according to the turnings conditions it faces. Moreover, they help to reduce the ambiguity between translation and rotation. Therefore the robot can deal with different turnings conditions and the navigation method will be more robust and accurate. Experimental results on challenging scenarios on a real ground robot demonstrate the effectiveness of sloped funnel lane technique. 
\end{abstract}

\section{Introduction}
\label{introduction}
The process of determining and following a safe and appropriate path from a starting point to a goal point is called navigation. There are various methods which use different sensors to perform it. Recently, visual navigation methods have been considered by the researchers due to the development of powerful processing modules and the expansion of their applications in mobile robots. These methods are used in both ground \cite{4,5,6,9,10,15,20,21} and flying \cite{11,12,13,14,19} autonomous robots. 

Regardless of the kind of robots, the visual navigation methods can be categorized into two types: map-based and map-less visual navigation\cite{1}. 

Map-based visual navigation methods \cite{18,20,21} are based on a model of the environment (map) where the robot has to find its location on it.

Map-less visual navigation methods do not need such a model to navigate in the environment \cite{16,17,22}. The robot depends on the elements observed in the environment to navigate. 

Some navigation methods display the environment with sequential images which characterize the desired path. They are considered as map-less visual navigation methods that are based on visual teach and repeat technique. The main advantages of these methods are scalability, not needing global metric map construction, and simple implementation. The images can be gathered easily from an environment. These methods can have more applications especially for robots with limited memory. 
On the other hand, according to the lack of scale and geometric information, following such paths is not an easy task. 

\begin{figure}[H]
\centering
\includegraphics[width=0.5\textwidth]{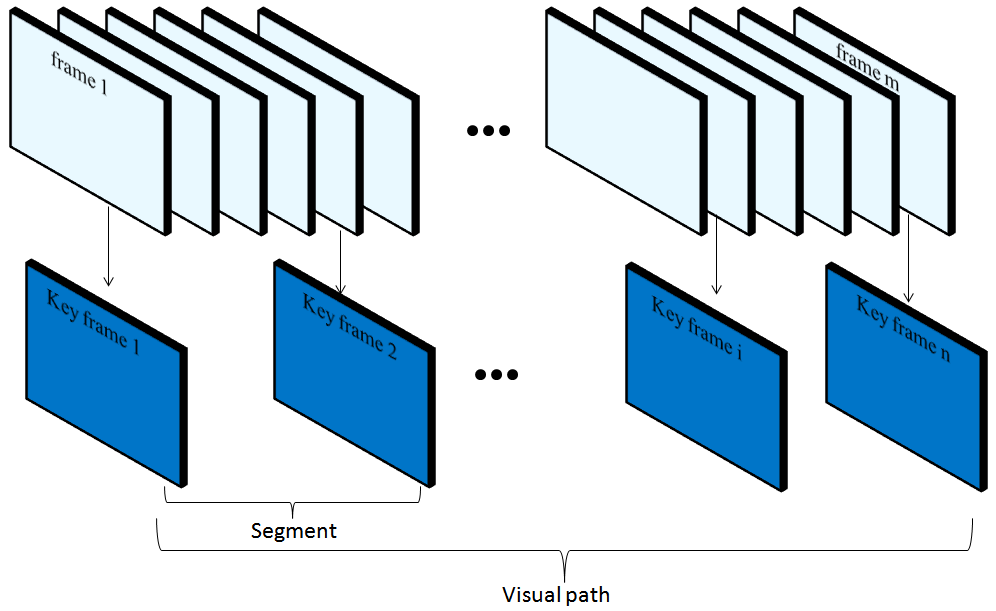}
\caption{Teaching phase (the keyframes are extracted from the recorded video}
\label{teachingphase} 
\end{figure}

In this paper, our navigation system falls into the category of visual teach and repeat technique. In the teaching phase (Fig. \ref{teachingphase}), the robot is guided to follow a path while recording a video, after that keyframes are extracted from the recorded video to make the visual path. The intervals between two consecutive keyframes are called segments.
\begin{figure}[H]
\centering
\includegraphics[width=0.5\textwidth]{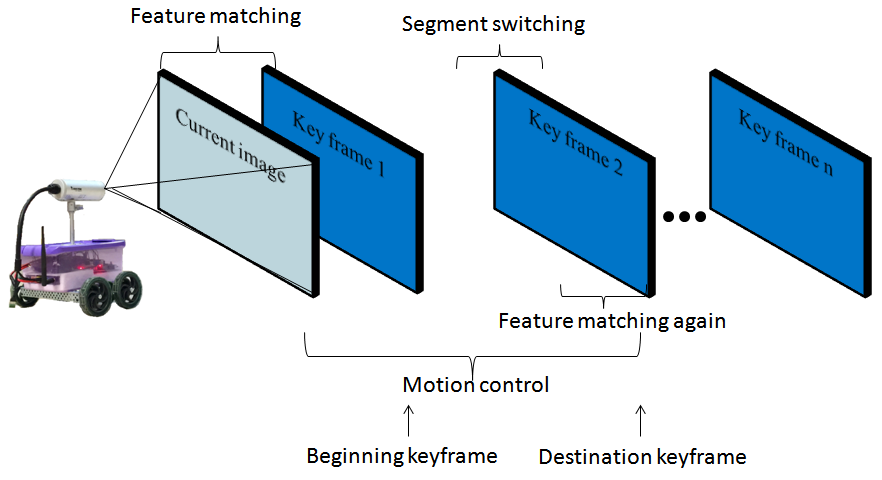}
\caption{Repeating phase }
\label{repeatingphase} 
\end{figure}
In the repeating phase (Fig. \ref{repeatingphase}), the robot has to be able to follow the visual path autonomously. Usually, a method is used to control the robot inside a segment in the visual path and a criterion is defined to switch from a segment to the next segment until reaching the last keyframe. Visual servoing is a well-known technique which is used to control the robot inside a segment. Visual servoing approaches usually need calculations such as Jacobian \cite{3} and homography or fundamental matrix \cite{10,11,12}. 

Another approach is the funnel lane that was proposed by Chen and Birchfield \cite{5}. The robot follows the path by making qualitative comparisons between the features extracted from the images in the teaching phase and the repeating phase. The method does not require any calculation to relate world coordinates to image coordinates.
Funnel lane assumes that the optical axis of the attached camera is parallel to the heading direction of the robot. For each feature, a region is determined based on two constraints of that feature. These regions are called funnel lanes. The intersection of these funnel lanes forms the combined funnel lane. The robot tries to keep itself inside the combined funnel lane to reach to its destination. Funnel lane has been implemented on ground robot \cite{5,6} and on quadrotor \cite{2,14,24}.

Standard funnel lane theory has its limitations. It specifies a region for the robot so that the robot can follow the visual path. The robot is controlled by getting left, right and straight moving commands. However, funnel lane does not have the ability to provide any information for the robot to know how much it should turn. For this reason, in funnel lane, the robot's radius of rotation is pre-set (translation and rotational speed of the robot are set beforehand \cite{5}) and the robot turns using the same radius along the whole path when turning is required. Therefore, the robot is not able to deal with all turning conditions. The robot can never deal with rotation in place which is important especially in narrow places because it will not move forward at all. This limitation should be taken into account in the teaching phase in order to be able to follow the visual path in the repeating phase. In other word, the robot's radius of rotation in the teaching phase should be set with regards to its value in the repeating phase or vice versa. As a result, the robot is not allowed to take all kinds of paths in the teaching phase as well. In addition, due to this limitation, the robot faces difficulty in correcting its path when it deviates from the desired path. This shortcoming decreases the robot maneuverability and limits the robot movements.

Another limitation is the occasional ambiguity between translation (forward movement) and rotation (turning movement) inside the funnel lane. This ambiguity can cause the robot to deviate from the desired path as we will explain later. This ambiguity was mentioned by the authors \cite{5} themselves however they tried to reduce this shortcoming by using odometry information. 

In this paper, we introduce \textbf{sloped funnel lane} which does not have these limitations. In sloped funnel lane, the robot is free to take any path with different turning conditions in the teaching phase. As well in the repeating phase, the robot sets the radius of rotation according to the situations it faces. The ambiguity is resolved without using any other sensors. Instead of creating a funnel lane for each feature and intersecting them to form the combined funnel lane, one funnel lane is created by looking at all features together. Also, two slopes based on the whole features are added in one step to the funnel lane. Therefore the proposed method is called sloped funnel lane. One of the slopes is used to determine the radius of rotation and help to reduce the ambiguity between translation and rotation. The other slope is used to keep the robot moving by a balance way throw the funnel lane. 

In the rest of this paper, first, some notations and assumptions are introduced which are used throughout the paper. After that, the method to create the visual path will be discussed. 
In section \ref{funnel lane}, we have a brief discussion about the funnel lane concept and its limitations. Then we will explain the sloped funnel lane which is proposed in this paper and we show how the sloped funnel lane overcomes the limitations of the standard funnel lane. After that, experimental examples that show how the proposed sloped funnel lane successfully follows a visual path in which the standard funnel lane failed to follow, is presented. Finally, we will have a conclusion.
\section{Notations and assumptions}
\label{Notations and assumptions}
In visual navigation systems some assumptions must be considered: enough light exists in the environment, the scene is often static, the environment contains enough texture to extract enough features, there is sufficient overlap between consecutive keyframes and the change of the conditions in the teaching phase and repeating phase does not affect the feature matching process in the repeating phase very much. \\
Some notations are used in this paper as follows:

\begin{itemize}
\item $c$ is the current image of the robot.

\item $V_i$ is the video taken from path $i$. 

\item $KF_{i,j}$ is the keyframe number $j$ in path $i$. 

\item $KFs_{i}$ is all keyframes in path $i$.

\item $S_{i,j}$ is the segment $j$ in path $i$, {$S_{i,j}: j \in \{1,2,…..n-1\}$}.

\item $F_{a}$ features of image $a$.

\item $RF_{a}$ right features of image $a$.

\item $LF_{a}$ left features of image $a$.

\item $MF(a,b)$ matched features of image $a$ with image $b$ (in image $a$).

\item $MF(b,a)$ matched features of image $b$ with image $a$ (in image $b$). Note that $MF(b,a)$ is different with $MF(a,b)$ because the coordinates of the matched features in image $a$ are not necessarily similar to the coordinates of the matched features in image $b$. 

\item $NMF(a,b)$ is the number of matched features of image $a$ with image $b$.

\item $\sigma_{MF(a,b)}$ is the standard deviation of $x$ coordinates of $MF(a,b)$.

\item $StdRatio(a,b)$ is the ratio of standard deviation of $x$ coordinates of $MF(a,b)$ to the standard deviation of $x$ coordinates of $MF(b,a)$.

\item $ED(a,b)$ is the Euclidean distance between the median of $x$ coordinates of $MF(a,b)$ and the median of $x$ coordinates of $MF(b,a)$.
\end{itemize}

\begin{figure}
\centering 
\includegraphics[width=0.5\textwidth]{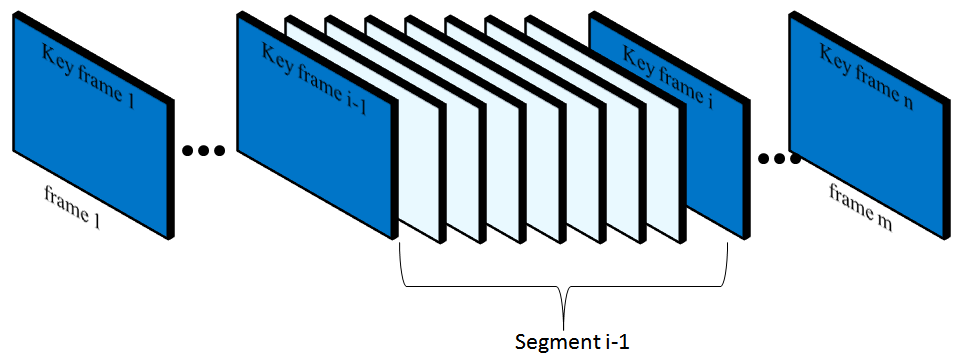}
\caption{$n$ keyframes are selected from $m$ frames to create a visual path and segment $i-1$ as shown is the interval between keyframe $i-1$ and keyframe $i$}
\label{visualpath} 
\end{figure}

Figure \ref{visualpath} shows a video recorded from a path consisting $m$ frames, $n$ selected keyframes and segment $i-1$, which is the interval between keyframe $i-1$ and keyframe $i$.
\section{visual path creation}
\label{visual path creation}
A robot is controlled to follow a path manually while it is recording a video. Some keyframes are selected from the video. The selected keyframes are called \textbf{visual path}. To select these keyframes, features of the first frame are detected and tracked in the video. A keyframe is selected when the percentage of successfully tracked features falls below 50 percent \cite{5}. The process is repeated until reaching the end of the video. The remaining successfully tracked features in each segment are stored with their coordinates because they are used in the repeating phase.

\section{Standard Funnel lane}
\label{funnel lane}
Standard funnel lane concept was introduced by Chen and Birchfield \cite{5}. The robot is controlled such that it is able to reach a destination image according to the image it receives from its attached camera. 
The camera optical axis is parallel to the robot heading and its optical axis passes through the axis of rotation of the robot. In the following, we explain the standard funnel lane. Then the motion control based on it will be described.

\begin{figure}
\centering
\includegraphics[width=0.5\textwidth]{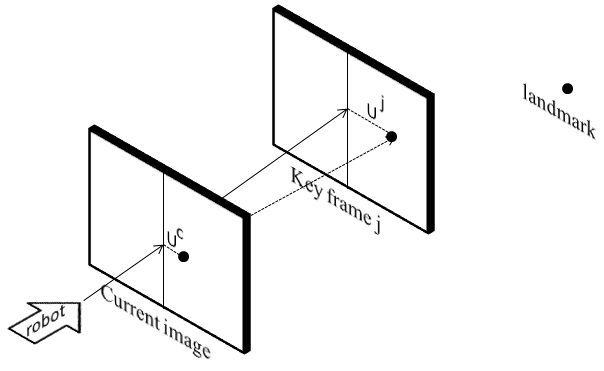}
\caption{A robot is moving on straight line with a camera attached on it which its optical axis is parallel to the robot's heading.}
\label{robotmoving} 
\end{figure}

Suppose that the robot wants to move from the current location to location $KF_{i,j}$. There are some fixed landmarks that are seen in the camera of the robot in both locations as shown in figure \ref{robotmoving}. Suppose we have both the current image and the destination keyframe image and the origin of the feature's coordinates is at the intersection of the optical axis and the image plane. If the robot goes forward in a straight line with the same heading direction as that of $KF_{i,j}$, the point $u^c$ will move away from the origin of the feature's coordinates toward $u^j $. When the robot reaches the destination, point $u^c$ will reach $u^j $. Therefore the funnel lane is defined as follows:

\textbf{Definition 1}: A funnel lane of a fixed landmark L and a robot location $KF_{i,j}$ is the set of locations $ FL_{L,KF_{i,j}}$ such that, for each $ C \in FL_{L,KF_{i,j}}$, the two funnel constraints are satisfied \cite{5}:
\begin{ceqn}
\begin{equation*}
|u^c |<|u^{j}| 
\end{equation*}
\begin{equation*} 
sign(u^c )= sign(u^{j}) 
\end{equation*}
\end{ceqn}
where $u^c$ and $u^{j}$ are the horizontal coordinates of the image projection of $L$ at locations $C$ and $KF_{i,j}$, respectively.

\begin{figure*}
\centering
\begin{subfigure}[b]{0.5\textwidth} 
\centering
\includegraphics[width=\linewidth]{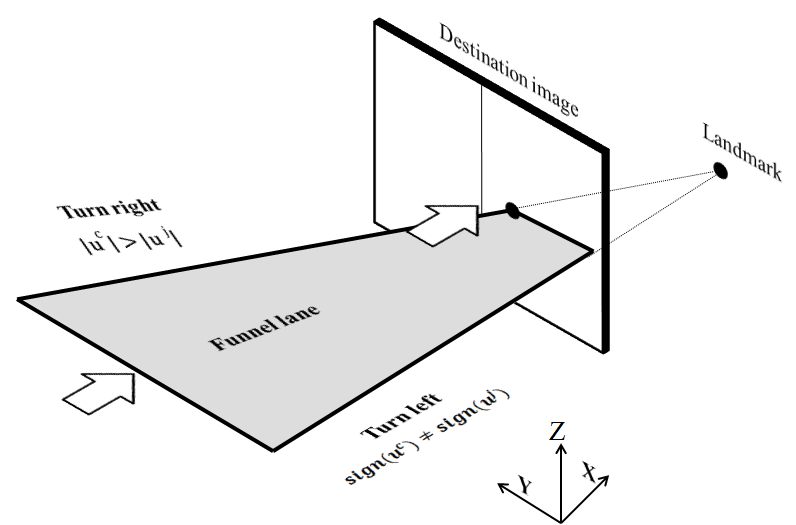}
\caption{}
\label{funnellaneOL}
\end{subfigure}\hfill
\begin{subfigure}[b]{0.5\textwidth} 
\centering
\includegraphics[width=\linewidth]{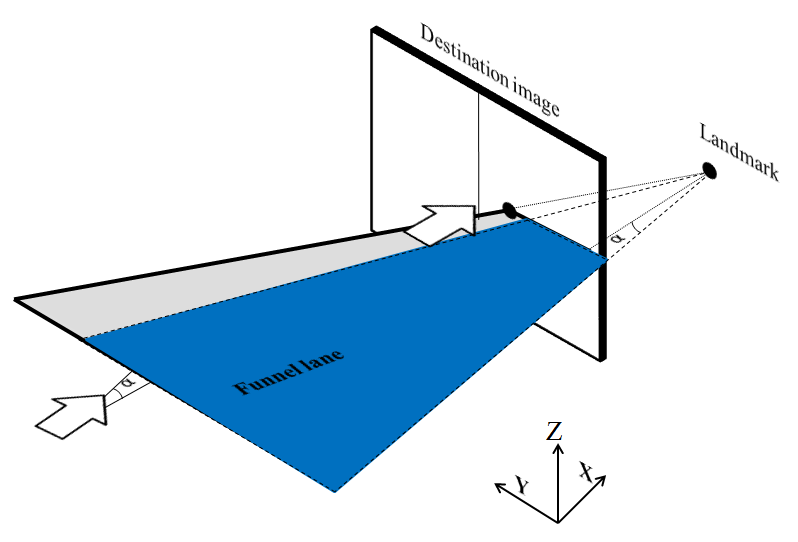}
\caption{}
\label{funnellanealpha}
\end{subfigure}\hfill
\caption{(a) Funnel lane created when the robot has the same heading angle with the destination, (b) Funnel lane created when the robot has a heading angle $\alpha$ with the destination}
\label{funnellane} 
\end{figure*}

If the robot is on the path toward the destination keyframe $KF_{i,j}$ with the same heading direction, the funnel lane will be as shown in figure \ref{funnellaneOL}. Note that the region is specified by two lines which represent the constraints of the funnel lane. The two constraints are satisfied when the robot is inside the funnel lane. For a right side feature ($u^j>0$), the first constraint ($|u^c |<|u^{j}|$) is violated when the robot exits from the right side and the second constraint ($sign(u^c )= sign(u^{j})$) is violated when it exists from the left side. For a left side feature ($u^j<0$) the opposite is true.

If the heading direction of the robot is not the same direction of the destination keyframe $KF_{i,j}$, the lines of the funnel lane are rotated by an angle depending on the angle that the robot has with destination keyframe $KF_{i,j}$ as shown in figure \ref{funnellanealpha}. 

\begin{figure}
\centering
\includegraphics[width=0.5\textwidth]{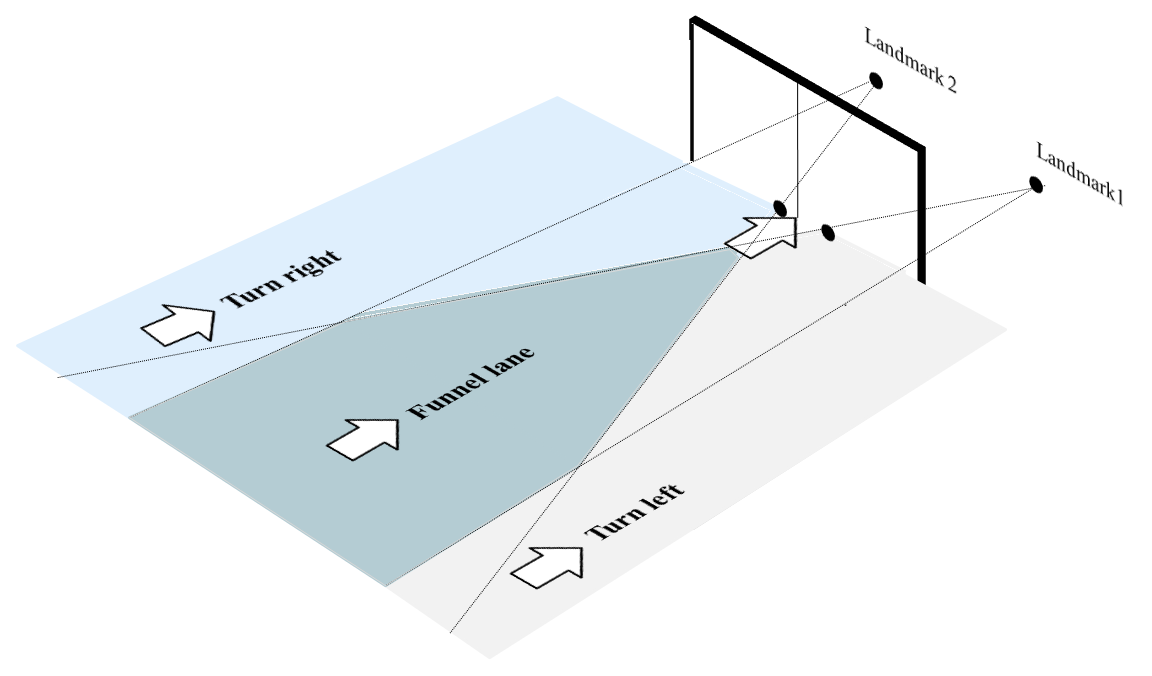}
\caption{Final funnel lane created by two features when the robot has the same heading angle with the destination}
\label{finalfunnellane} 
\end{figure}

For each landmark, a funnel lane region is created. By intersecting all funnel lanes, a combined funnel lane is obtained in which the constraints of all features are satisfied. Figure \ref{finalfunnellane} shows an example of how the combined funnel lane will be if we have two features. 

\subsection{Motion control based on standard funnel lane}
\label{motion control}

First, the features of the current image are matched with the features of the beginning $KF_{i,j-1}$ in the segment. Then, the matched features are tracked and their horizontal coordinates are compared with the horizontal coordinates of their correspondence features in the destination $KF_{i,j}$. If no constraint for each feature is violated the robot continually moves forward because it is assumed to be inside the combined funnel lane. Whenever constraint 1 of a right side keyframe feature ($u^j>0$) is violated it means that the robot has gone outside the funnel lane from the left side so it has to get a right turning command and whenever constraint 2 of a right side feature is violated it means that the robot has gone outside the funnel lane from the right side so it has to get a left turning command to get it back to the funnel lane. If the keyframe feature is left side ($u^j<0$) the directions are reversed. The constraints are checked for each feature. The final command will be the majority command gets by all features.

\subsection{Limitations}
\label{funnel lane limitations}
Motion control based on standard funnel lane has some limitations which are:

1- Constant radius of rotation 

In funnel lane, the robot is moving forward and it turns by an amount to the right or to the left depending on the command it gets\cite{5,6}. Note that the translational and rotational speeds are set beforehand. In another word, the radius of rotation of the robot is set beforehand. This reduces the maneuverability of the robot. The robot cannot take any path in the teaching phase. Moreover in the repeating phase according to this reduction of maneuverability the robot cannot correct its direction easily when it deviates from the desired path especially in turnings. 

2- The ambiguity of translation and rotation

\begin{figure}
\centering
\begin{subfigure}[b]{0.4\textwidth}
\centering
\includegraphics[width=\linewidth]{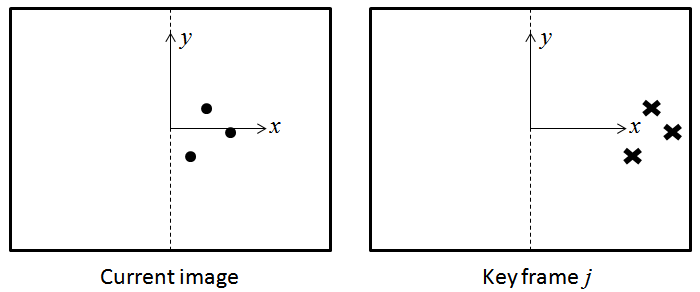}
\caption{}
\label{featurebothsides1}
\end{subfigure}\hfill
\begin{subfigure}[b]{0.4\textwidth}
\centering
\includegraphics[width=\linewidth]{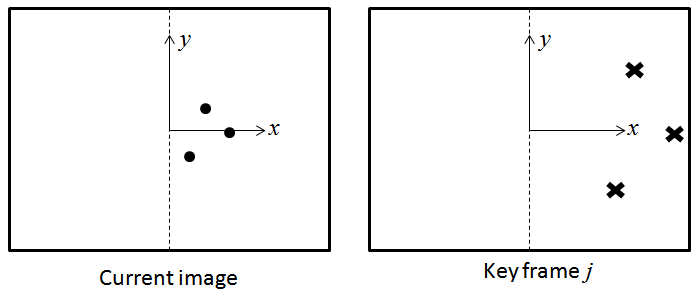}
\caption{}
\label{featurebothsides2}
\end{subfigure}\hfill
\caption{Circle symbols show the positions of the current features and star symbols show the positions of their corresponding features in destination keyframe $j$. (a) A left turning causes the destination features to lay on the right side of the current features, (b) A forward movement causes the destination features to lay on the right side of the current features}
\label{featurebothsides} 
\end{figure}

An ambiguity exists between translation (going strai\-ght) and rotation (turning) inside the funnel lane itself \cite{5}. Falling inside the funnel lane does not necessarily mean a translation command to the robot. To make it more clear consider figure \ref{featurebothsides} where there are features just in the right side and the $x$ coordinates of the destination features lay on the right side of the current features. In the first case, a turning causes the destination features lay on the right side of the current features (figure \ref{featurebothsides1}). In the second case the path is straight forward and therefore the destination features lay on the right side of the current features (figure \ref{featurebothsides2}). In the standard funnel lane, the two constraints ($|u^c|<|u^j|$ and $sign(u^c)==sign(u^j)$) are satisfied for all features and the robot falls on the combined funnel lane, which means it will get a straight forward command for both cases. This causes the robot to deviate from the desired path in case of figure \ref{featurebothsides1}.

\begin{figure}
\centering 
\begin{subfigure}[b]{0.4\textwidth}
\centering 
\includegraphics[width=\linewidth]{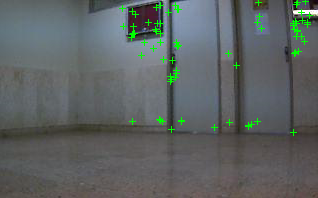}
\caption{}
\label{features in keyframe1}
\end{subfigure}\hfill
\begin{subfigure}[b]{0.4\textwidth}
\centering 
\includegraphics[width=\linewidth]{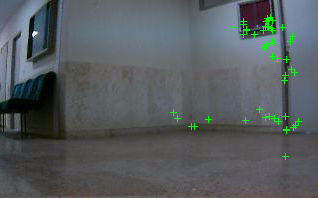}
\caption{}
\label{features in keyframe2}
\end{subfigure}\hfill
\caption{Two consecutive keyframes selected in left turning condition, (a) the first keyframe and (b) the next keyframe.}
\label{shiftedfeatures} 
\end{figure}

Existing destination features on both sides of the image help to narrowly constrain the path of the robot. This explains why existing features on both sides in standard funnel lane is necessary \cite{5}. But unfortunately, the ambiguity will remain inside the funnel lane. Moreover, it is not guaranteed that the destination matched features lay on both sides. In turning conditions the tracked features come out from the frame and the remaining common features between two consecutive key\-frames will be shifted to the right or to the left side of the image. In other words, the common features in the destination keyframe will be shifted. To make it clear consider figure \ref{shiftedfeatures} which shows two consecutive keyframes which are selected to create the visual path in turning condition. As it is seen the remaining features are shifted to the right because a turning to the left has occurred. In addition in the repeating phase at the feature matching process, not all features are matched due to changes of view, light, etc. Moreover, some features are lost due to tracking failure (inside the segment) or due to moving objects. As a result especially in turning conditions the destination matched features are not guaranteed to be on both sides. 

3- No control inside the funnel lane

The robot is moving forward until it gets out from the funnel lane. After getting out it receives a command to return it back to the funnel lane. 

\section{Sloped funnel lane}
\label{sloped funnel lane}
Sloped funnel lane is a method which overcomes the shortcomings of the standard funnel lane. First, we will explain the sloped funnel lane. Then the motion control based on it will be described. After that, we will show how the sloped funnel lane can overcome the limitations of the standard funnel lane.

The standard funnel lane gives no information about the radius of rotation, and there is an ambiguity between translation and rotation as explained. The standard funnel lane is created according to the fact that the features will move away from the center of the feature's coordinates toward the edge of the images when the robot moves in a straight line toward the destination image.

Actually, in standard funnel lane for each feature, a funnel lane is created and later they are combined. However, more information can be extracted by looking at all features together. 
In straight movements, as seen from the robot's camera features move away from the center, in addition, will move away from each other as the robot moves forward.
So, we can conclude that the ratio of the standard deviation of $x$ coordinates of all matched features in the current image to the standard deviation of $x$ coordinates of their corresponding features in destination image will become greater as the robot moves forward toward the destination. 

To take this fact into account, we add slopes to the standard funnel lane. The idea is inspired by the movement of a ball on a sloped surface. If the surface has a slope toward front, the ball moves forward. If the surface has a slope toward left or right sides the ball will roll to the left or right. Moreover, if the surface has a slope toward front and left \slash right side at the same time the ball will roll forward and tend to the left \slash right. Depending on the amount of the slope toward forward and toward left \slash right the ball will roll in different trajectories. 

\begin{figure}
\centering 
\includegraphics[width=\linewidth]{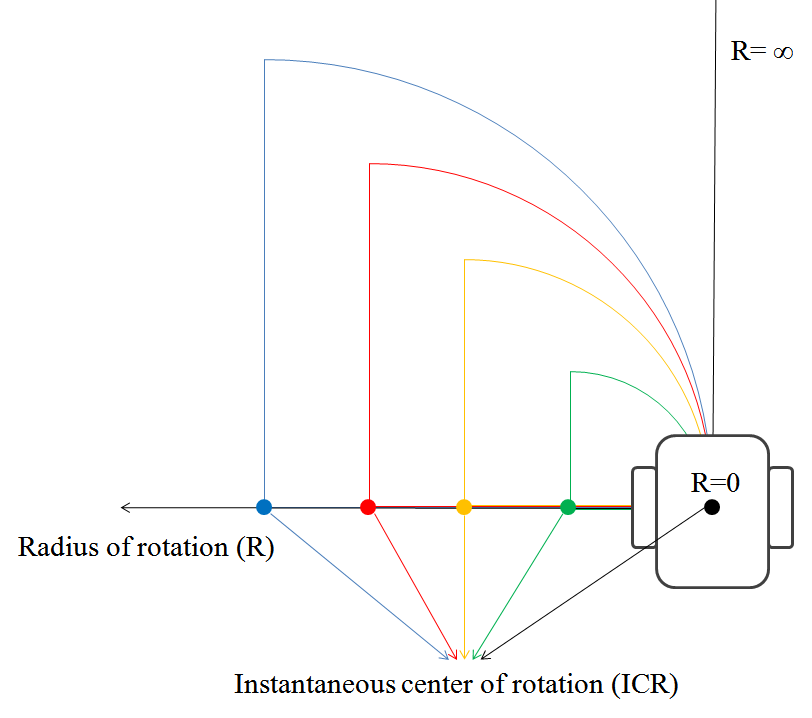}
\caption{Different trajectories with different radiuses of rotations are shown when the robot turns to the left} 
\label{leftTurnings}
\end{figure}

In our case, the ball is the robot and the surface is the sloped funnel lane. The different trajectories are considered to be turnings with different radii of rotations. In figure \ref{leftTurnings} different trajectories with different radii of rotation when the robot turns to the left are shown. 
To simplify things, the radius of rotation is specified through the forward slope. The sharper slope means the larger radius of rotation. The right and left slopes are only used to determine the direction of the turn or whether the robot should turn or not.
In a nutshell, if there is a right or left slope, the robot will turn right or left according to the radius of turning specified by the forward slope, otherwise the robot will not turn.

To define such a surface we define the slope around $y$ axis inversely proportional to the ratio of the standard deviation and the slope around $x$ axis is defined proportional to the difference of current and destination feature coordinates. 

The farther the current image is from the destination keyframe, the slope of the funnel lane around $y$ axis should be larger, and it is reduced when we go toward the destination keyframe. Thus we define this slope inversely proportional to the ratio of $\sigma_{MF(c,KF_{i,j})}$ to $\sigma_{MF(KF_{i,j},c)}$:
\begin{ceqn}
\begin{equation} S_y=1-\dfrac {\sigma_{MF(c,KF_{i,j})}}{\sigma_{MF(KF_{i,j},c)}} \end{equation}
\end{ceqn}
In addition, the slope around the $x$ axis depends on the distance of the current features with the destination features. The more difference causes the more slope. This slope is used to control the robot inside the funnel lane. We calculate two slopes according to the right and left features. The features of the current image are considered as right or left features according to being on the right or left side of the destination keyframe. Two features that represent right and left features are chosen. The feature that represents the right features is the median of the right features ($\mu_r$) and the other one that represents the left features is the median of left features ($\mu_l$). In case of existing just one feature at each side, the only existing feature is chosen to represent the side. In the absence of the right or left features, the sloped is created just by one feature that represents the other ones. The right features create a negative slope around the $x$ axis while the left features create a positive slope. The final slope is the sum of both slopes. It is noteworthy that, the slopes should be normalized before summing their values in order to balance between left and right features. So we define the slope around $x$:
\begin{ceqn}
\begin{equation} S_x=\dfrac{\mu_l^c - \mu_l^j }{|\mu_l^j|} + \dfrac{\mu_r^c - \mu_r^j}{|\mu_r^j|} \end{equation} 
\end{ceqn}
where $\mu_l^{j}$ and $\mu_l^c$ are the median coordinates of the left features at the location $KF_{i,j}$ and the median coordinates of their correspondences at location $c$, respectively. $\mu_r^{j}$ and $\mu_r^c$ are the median coordinates of the right features at the location $KF_{i,j}$ and the median coordinates of their correspondences at location $c$, respectively. 

\begin{figure}
\centering
\begin{subfigure}[b]{0.24\textwidth}
\centering 
\includegraphics[width=\linewidth]{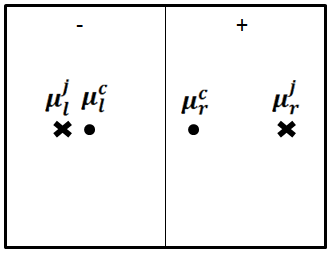}
\caption{} 
\label{positiveSlope}
\end{subfigure}\hfill
\begin{subfigure}[b]{0.24\textwidth} 
\centering
\includegraphics[width=\linewidth]{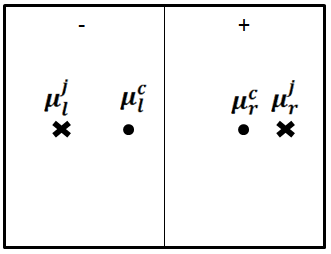}
\caption{} 
\label{negativeSlope}
\end{subfigure}\hfill
\caption{Two examples of summing the slope of both sloped funnel lanes representing each side: (a) the sum will be positive, (b) the sum will be negative. }
\label{summingSlopes} 
\end{figure}

Figure \ref{summingSlopes} shows an example of summing these two slopes. The sum of two slopes in figure \ref{positiveSlope} will be positive and in figure \ref{negativeSlope} will be negative. 

This slope is used to control the robot inside the funnel lane itself. Instead of waiting for the robot to get out from the funnel lane, this slope helps to keep the robot inside it. 
These two slopes are added to the funnel lane and as we mentioned in sloped funnel lane just one funnel lane is created by all features together. Therefore the definition of the sloped funnel lane will be as the following:

\textbf{Definition 2}: A sloped funnel lane (SFL) of a set of fixed landmarks L, where some of them are left landmarks $L_l:1$ to $m$ (projected on the left side of the destination keyframe) and the others are right landmarks $L_r:m$ to $n$ (projected on the right side of the destination keyframe) at a robot location ($KF_{i,j}$) is the set of locations $SFL_{L,KF_{i,j} }$ such that, for each $C \in SFL_{L,KF_{i,j} }$, the following four funnel constraints are satisfied:
\begin{ceqn}
\begin{equation*}
|\mu_r^c |<|\mu_r^{j}| 
\end{equation*}
\begin{equation*}
|\mu_l^c |<|\mu_l^{j}| 
\end{equation*}
\begin{equation*}
sign(\mu_r^c )= sign(\mu_r^{j} ) 
\end{equation*}
\begin{equation*}
sign(\mu_l^c )= sign(\mu_l^{j} ) 
\end{equation*}

and the funnel lane slope around $y$ axis (pitch) is:
\begin{equation*} S_y=1-\dfrac {\sigma_{MF(c,KF_{i,j})}}{\sigma_{MF(KF_{i,j},c)}} \end{equation*}

and the slope around $x$ axis (roll) is:
\begin{equation*} S_x=\dfrac{\mu_l^c - \mu_l^j }{|\mu_l^j|} + \dfrac{\mu_r^c - \mu_r^j}{|\mu_r^j|} \end{equation*}
\end{ceqn}
where $\mu_l^{j}$ and $\mu_l^c$ are the median coordinates of the image projection of $L_l:1-m$ at the location $KF_{i,j}$ and the median coordinates of their correspondences at location $c$, respectively. $\mu_r^{j}$ and $\mu_r^c$ are the median coordinates of the image projection of $L_r:m-n$ at the location $KF_{i,j}$ and the median coordinates of their correspondences at location $c$, respectively. $\sigma_{MF(c,KF_{i,j})}$ and $\sigma_{MF(KF_{i,j},c)}$ are the standard deviation of the coordinates of the matched features of current image with the destination keyframe $KF_{i,j}$ at locations $c$ and $KF_{i,j}$, respectively.

\begin{figure*}
\centering
\begin{subfigure}[b]{0.5\textwidth}
\centering
\includegraphics[width=\linewidth]{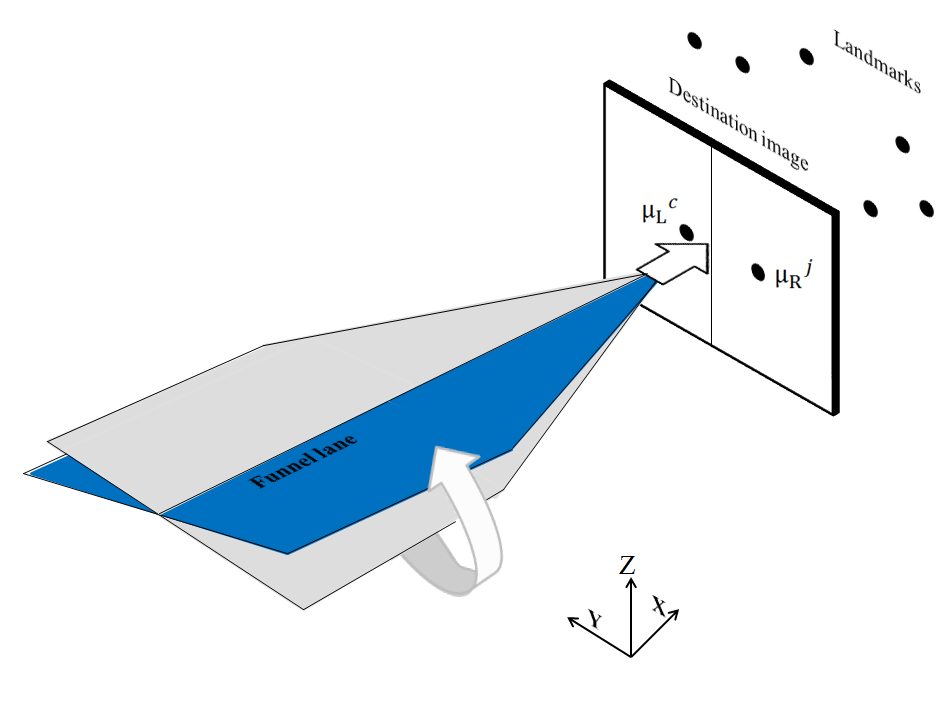}
\caption{}
\label{finalfunnellaneRx}
\end{subfigure}\hfill
\begin{subfigure}[b]{0.5\textwidth}
\centering
\includegraphics[width=\linewidth]{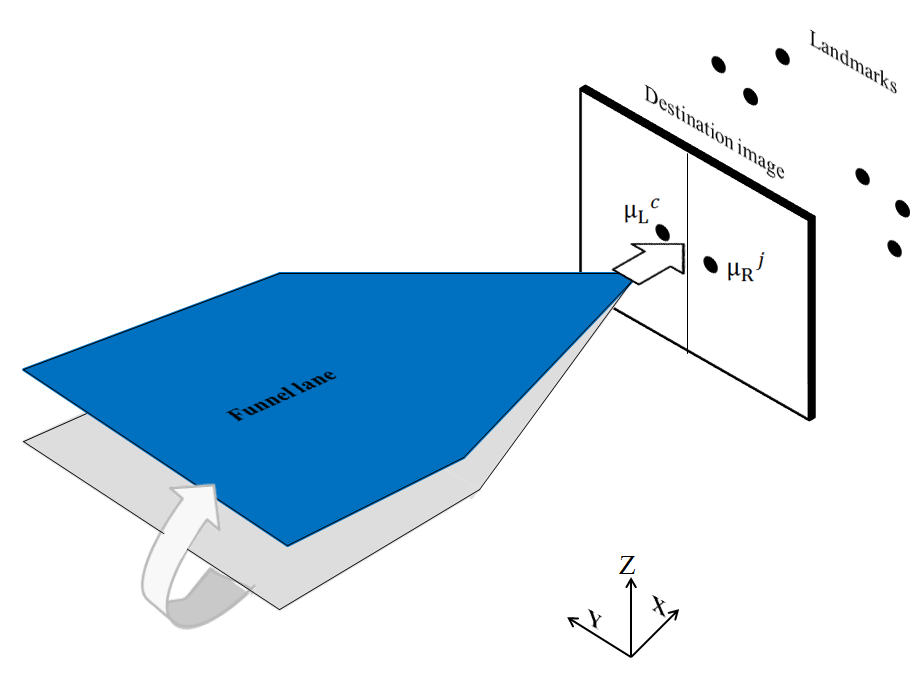}
\caption{}
\label{finalfunnellaneRy}
\end{subfigure}\hfill
\begin{subfigure}[b]{0.5\textwidth}
\centering
\includegraphics[width=\linewidth]{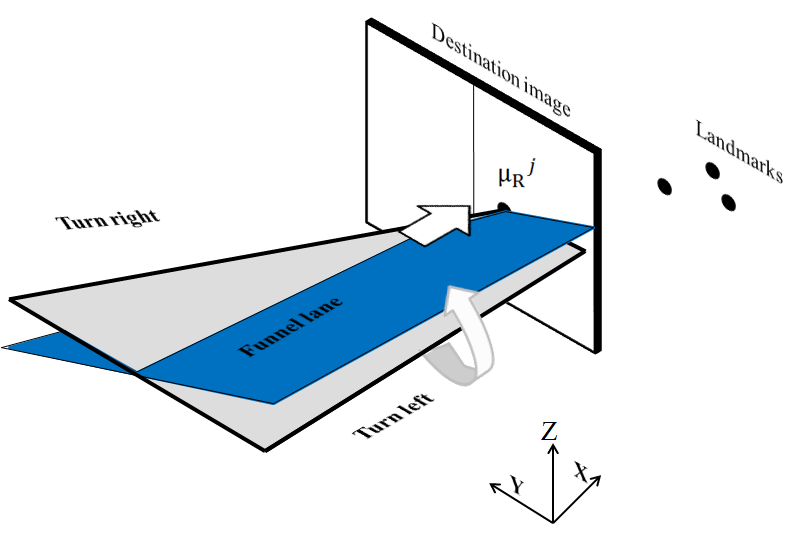}
\caption{}
\label{funnellaneRx}
\end{subfigure}\hfill
\begin{subfigure}[b]{0.5\textwidth}
\centering
\includegraphics[width=\linewidth]{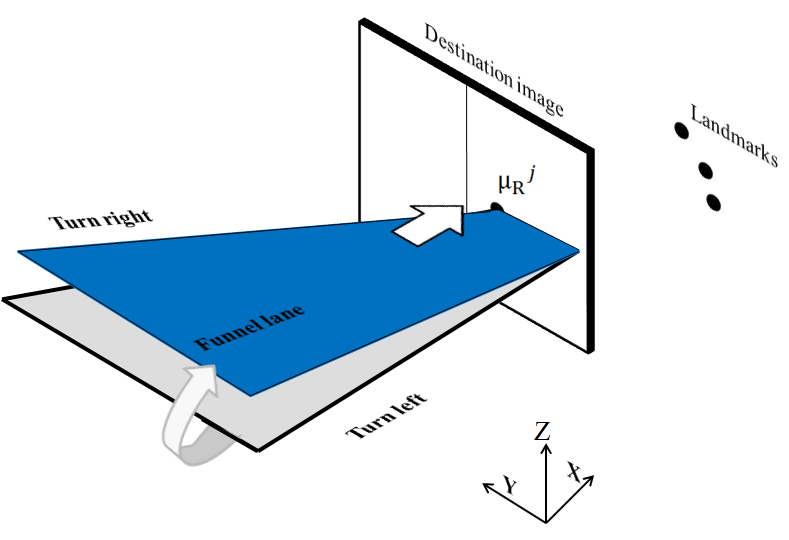}
\caption{}
\label{funnellaneRy}
\end{subfigure}\hfill
\caption{(a) Sloped funnel lane created with a negative slope around $x$ axis (roll), (b) Sloped funnel lane created with slope around $y$ axis (pitch). The obtained sloped funnel lane in case of the absence of the left features is shown in (c) with a negative slope around $x$ axis and in (d) with slope around $y$ axis (pitch)}
\label{slopedfunnellane} 
\end{figure*}

Figure \ref{finalfunnellaneRy} shows the obtained sloped funnel lane when the robot heading angle is the same as the destination keyframe with a slope around the $y$ axis and no slope around the $x$ axis ($S_y>0$ and $S_x=0$ which means a forward movement should happen). Figure \ref{finalfunnellaneRx} demonstrates with the same conditions but with just a negative slope around the $x$ axis ($S_y=0$ and $S_x$ is negative which means a left turning in place should happen).
Figure \ref{funnellaneRx} shows a sloped funnel lane with a slope around the $y$ axis (pitch) and figure \ref{funnellaneRy} shows a sloped funnel lane with a slope around the $x$ axis (roll) in case of the absence of the left features. 

In the sloped funnel lane similar to the standard funnel lane if the heading direction of the robot is not in the same direction of the destination keyframe $j$, the lines of the funnel lane are rotated by an angle which is equal to the angle that the robot has with the destination keyframe $j$. 

\subsection{Motion control based on sloped funnel lane}
\label{motion control in sloped funnel lane}
The robot moves forward until it is inside a funnel lane with no slope around the $x$ axis. The robot is inside the funnel lane when the four constraints are satisfied. Whenever constraint 1 or constraint 4 are violated it means that the robot has gone outside the funnel lane from the left side so it gets a right turning command and whenever constraint 2 or constraint 3 is violated it means that the robot has gone outside the funnel lane from the right side so it gets a left turning command to keep it in the funnel lane. While the robot is inside the funnel lane but the funnel lane has a positive slope around the $x$ axis, the robot gets a right command and when it has a negative slope, it gets a left command. 
Note that the radius of rotation is determined according to the slope around the $y$ axis in all turnings commands. The less the $y$ slope, the sharper the robot turns and vice versa. As the slope around $y$ axis gets near zero, the radius of rotation in turning command will also be near zero and the turning will be more like rotation in place. 

The motion control based on the sloped funnel lane is presented in algorithm \ref{alg:motion control}.
\begin{algorithm}
\caption{Motion control based on sloped funnel lane }
\label{alg:motion control}
\begin{algorithmic}[1]
\State $Radius\!\ of\!\ rotation=f(S_y)$ 
\If {four constraints are satisfied}\Comment{inside SFL}
\If {$S_x ==0 $}\Comment{zero roll}
\State Move forward
\ElsIf { $S_x<0$ }\Comment{roll is negative}
\State Turn left
\ElsIf {$S_x>0$} \Comment{roll is positive}
\State Turn right 
\EndIf 
\Else
\If {constraint 1 or constraint 4 are violated} 
\State Turn right 
\ElsIf {constraint 2 or constraint 3 are violated}
\State Turn left
\EndIf
\EndIf
\end{algorithmic}
\end{algorithm}

\subsection{How sloped funnel lane does not have the limitations of standard funnel lane}
\label{overlimitations}
The sloped funnel lane can deal with the limitations that are mentioned in section \ref{funnel lane limitations}. We will demonstrate the limitations and explain how sloped funnel lane can handle them.

1- Constant radius of rotation

\begin{figure}
\centering
\begin{subfigure}[b]{0.4\textwidth} 
\centering
\includegraphics[width=\linewidth]{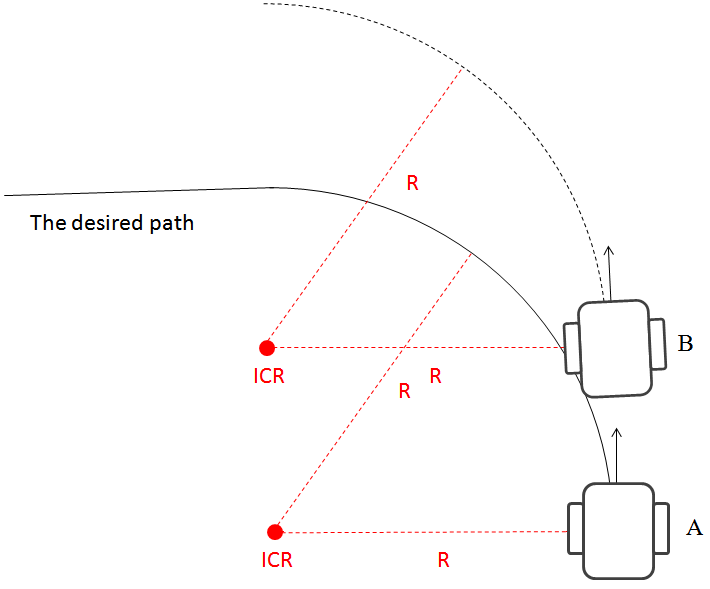}
\caption{}
\label{cr}
\end{subfigure}\hfill
\begin{subfigure}[b]{0.4\textwidth} 
\centering
\includegraphics[width=\linewidth]{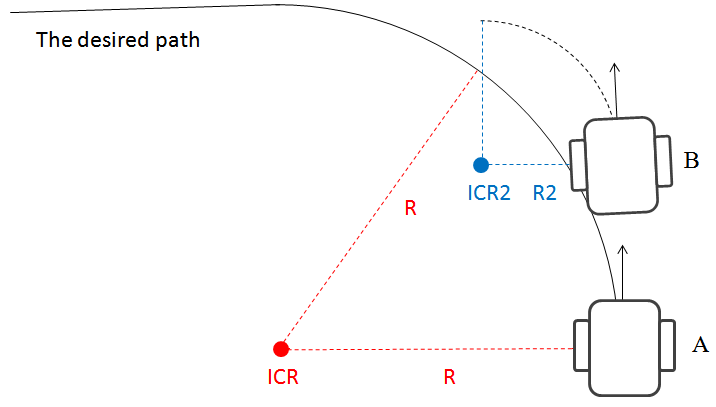}
\caption{}
\label{vr}
\end{subfigure}\hfill
\caption{}
\label{crandvr} 
\end{figure}

The radius of rotation is defined in the sloped funnel lane. As we explained, the slope around the $y$ axis determines the radius of rotation, which means that the robot has more maneuverability. It is free to take any path in the teaching phase with different turning conditions including rotation in place. In the repeating phase, the robot will set its radius of rotation adaptively, depending on the situation it faces. In addition, if the robot deviates from the path especially in turnings, it can correct its direction more easily by changing its radius of rotation. For example, as shown in figure \ref{crandvr}, suppose that the robot starts to follow the desired path from A. The robot in position B gets the turning command. In figure \ref{cr} the robot faces a problem to correct its direction due to its constant radius of rotation, while in figure \ref{vr} the robot corrects its direction easily. 

2- The ambiguity of translation and rotation

In the sloped funnel lane, a slope around the $y$ axis is added which looks at all features together. This slope helps to resolve the ambiguity of rotation and translation. A small slope means a small radius of rotation which means a small translation the robot has to do and vice versa. For example, in figure \ref{featurebothsides}, the standard funnel lane does not distinguish between both keyframes as we have shown before, but the slope around the $y$ axis in the sloped funnel lane helps to distinguish between them.

The reason is that the slope around the $y$ axis is inversely proportional to the standard deviation ratio which in the first case is closer to 1 than the second case. In both cases, a left command is sent. Therefore no features exist on the left side and slope around the $x$ axis will be negative. But in the first case, the robot will turn sharper near to rotation in place (less translation), and in the second case, a turning near to moving straight forward occurs (less rotation).

As a result, the sloped funnel lane by resolving this ambiguity prevents the robot from deviating and from getting out of the desired path. 

3- No control inside the funnel lane

In the sloped funnel lane, the slope around the $x$ axis is added. This slope is used to control the robot inside the sloped funnel lane. The slope helps the robot to move in a balanced way throw the funnel lane. This helps to keep the robot inside the funnel lane instead of waiting for leaving out of it.  

\section{keyframe switching criterion}
\label{keyframe switching criterion}
\begin{figure}
\centering
\includegraphics[width=0.4\textwidth]{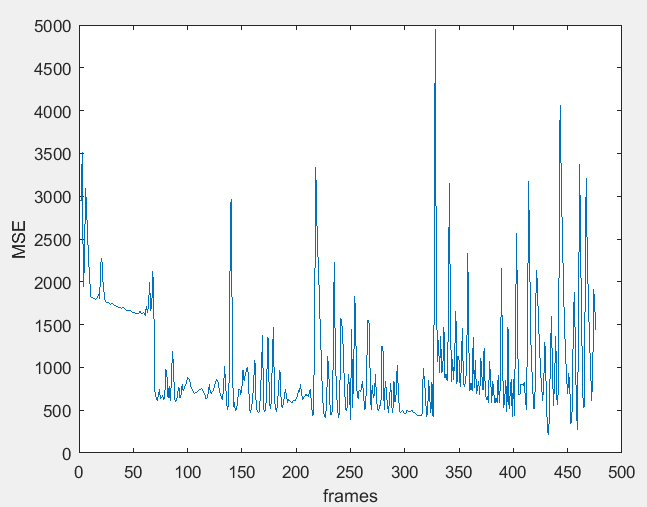}
\caption{A sample of MSE error in a real experiment}
\label{mse} 
\end{figure}

Funnel lane is a method to control a robot between two keyframes and how to move inside a segment. An important issue is how to define the criterion to switch to another keyframe. Mean square error between the coordinates of current features and features in the destination keyframe ($MSE_{c,j}$) can be used as a criterion. Chen and Birchfield \cite{6} proposed a method based on MSE. They supposed that the MSE error will become smaller as the robot moves toward the destination image, and the error is decreasing until reaching it. In practice, in our experiments, we noticed that this error was not decreasing uniformly due to losing features and insensitive steering. This criterion is related to the movement of the robot which makes it so sensitive. Figure \ref{mse} shows a sample of this error in a real experiment. As it is shown, the error was oscillating and a lot of switching happens because the criterion needs very sensitive steering. So steering a little more than necessary or even losing some features causes the MSE not to decrease. 
Another method uses mean square error with odometry information to define a probability for switching \cite{5}. We prefer to define a criterion just based on the features themselves, and not using odometry. In \cite{2} the switching is based on matching two successive keyframes \cite{2}. The features of the current image are matched with the features in the destination keyframe and with the features in the keyframe next to the destination keyframe. A switching happens whenever the number of matched features with destination keyframe becomes less than the number of matched features of the keyframe next to destination keyframe. Therefore two matchings are required for every cycle to know when to switch. 

In our work, a simple method based on the slope around $y$ defined in the sloped funnel lane is used. When StdRatio(current image, destination keyframe) becomes greater than 1 and the Euclidean distance of the median of both coordinates ED(current image, destination keyframe) becomes less than a threshold, a switching happens. 

\section{experimental results}
\label{sec:8}
\begin{figure}
\centering
\includegraphics[width=0.4\textwidth]{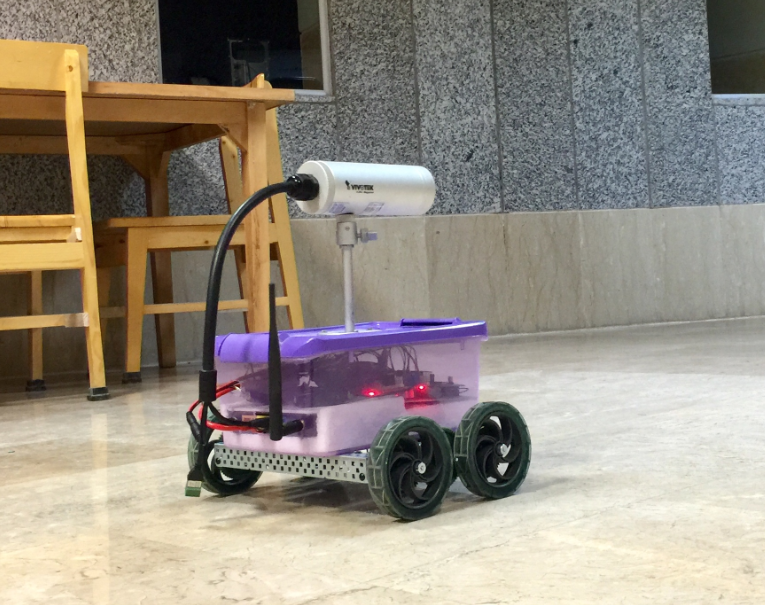}
\caption{The robot which is used to evaluate the proposed navigation method}
\label{robot} 
\end{figure}

Real experiments were conducted on a robot with a VEX platform \cite{23}. The robot uses an IP camera and sends the images $320 \times 200$ using WIFI to a laptop. Blob features are used in this paper. A well-known blob detection technique is SIFT \cite{8} that uses the difference of Gaussian operator to detect features. SURF \cite{7} is a speeded-up version of SIFT. It approximates the Gaussian with a box filter and the convolution with a box filter can be calculated simultaneously for different scales. In our experiments, we choose SURF detectors to speed up the navigation algorithm and its length is chosen to be 64. Larger length gives more accuracy but it decreases the speed of features matching. For feature tracking Kanade–Lucas–Tomasi (KLT) algorithm with default block size [31 31] is used.
The algorithm is executed on a laptop and the commands are sent to the robot for path following. The algorithm is implemented in MATLAB 2016 on a VAIO laptop (core i7 1.73GHz RAM 4GB). The robot is shown in figure \ref{robot}. First, the robot is controlled manually from the laptop while recording a video from the traversed path. After that, the visual path is constructed as explained in the previous sections. Then, the robot is placed on the same initial point and is controlled by the algorithm running on the laptop to follow the recorded visual path. 

The method used for visual navigation after creating the visual path is presented in algorithm \ref{alg:visual navigation}. 

\begin{algorithm}
\caption{visual navigation}
\label{alg:visual navigation}
\begin{algorithmic}[1]
\State assumed: The visual path i consists from n keyframes, robot starts from segment 1 
\State C=capture the current image
\State j=1
\State Detect surf features of $C$
\State Match features of $C$ with $KF_{i,j}$ 
\State $switch=false$
\State $NofF=NMF(C,KF_{i,j})$ 
\State {lost=false}
\While{$j<n$ or lost=false}
\If{$StdRatio(C,KF_{i,j+1})>1$ and $ED(C,KF_{i,j+1})<Threshold1$ or $switch=true$} \Comment {A switching to the next segment is happens}
\State $j=j+1$
\State C=capture the current image
\State Detect surf features of $C$
\State Match features of C with $KF_{i,j}$ 
\State $NofF=NMF(C,KF_{i,j})$
\Else \Comment {Control inside a segment}
\If {$NofF >Threshold2$ } \Comment {Sufficient features remained}
\State Track the matched features with KLT 
\State $NofF=NofF - lost features$
\State Control the robot with the sloped funnel lane
\Else
\State $time=0$
\While{ $NofF<Threshold2$} \Comment {Robot deviates or features lost}
\State C=capture the current image
\State Detect surf features of $C$
\State Match features of $C$ with $KF_{i,j}$ 
\State $NofF=NMF(C,KF_{i,j})$
\State {Stop the robot} 
\State {time=time+1;}
\If { $time>Threshold3$}
\State {lost=true}
\State {return}
\EndIf
\EndWhile 
\EndIf
\EndIf
\EndWhile
\State {Stop the robot}
\end{algorithmic}
\end{algorithm}
\begin{figure}
\centering
\begin{subfigure}[b]{0.4\textwidth} 
\centering
\includegraphics[width=\linewidth]{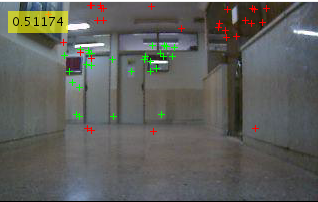}
\caption{}
\label{vp1}
\end{subfigure}\hfill
\caption{The matched features of the current image with the destination keyframe is shown by green color and their corresponding destination features are shown by red color, StdRatio(current image, destination keyframe) is shown at the top of the figure}
\label{matchedfeatures} 
\end{figure}
In section \ref{overlimitations} we show how the sloped funnel lane outperforms the standard funnel lane. In the sloped funnel lane unlike the standard funnel lane the robot is free to take any path (with different radius of rotations) in the teaching phase. 

Therefore, these experiments have been performed to show the impact of these restrictions on following the paths in the repeating phases even when the robot takes a path with a similar constant radius of rotation in the teaching phase. 

Six practical scenarios are considered to show that. Moreover, two paths are chosen to compare the accuracy and the repeatability of our method with the standard funnel lane. 

First, the visual path is created. Then the robot is placed at the initial point and it tries to follow the visual path once with the sloped funnel lane and again with the standard funnel lane. Figure \ref{matchedfeatures} shows the features in the current image and their correspondence features in the destination keyframe. Also $StdRatio(c,KF_{i,1})$ is shown at the top of the figure.

\subsection{Six practical scenarios}
\begin{figure}
\centering
\begin{subfigure}[b]{0.4\textwidth}
\centering 
\includegraphics[width=\linewidth]{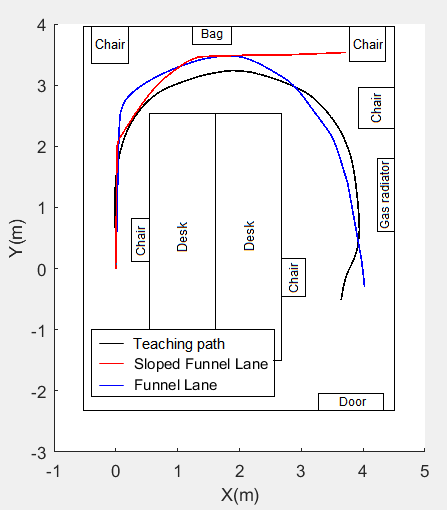}
\caption{} 
\label{insideroom}
\end{subfigure}\hfill
\begin{subfigure}[b]{0.4\textwidth} 
\centering
\includegraphics[width=\linewidth]{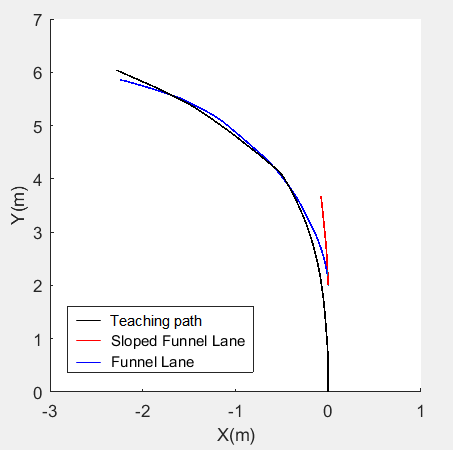}
\caption{} 
\label{inside2}
\end{subfigure}\hfill
\begin{subfigure}[b]{0.4\textwidth} 
\centering
\includegraphics[width=\linewidth]{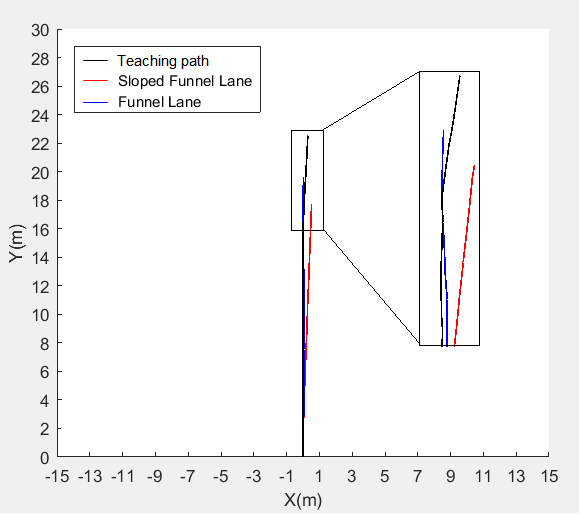}
\caption{} 
\label{inside3}
\end{subfigure}\hfill
\caption{(a) The first indoor scenario with narrow space (funnel lane failed) (b)The second scenario with a different initial point at the repeating phase (funnel lane failed) and (c) The third indoor scenario with an almost straight path. }
\label{insidescenarios} 
\end{figure}

The goal is to evaluate the path following the ability of both algorithms in six challenging scenarios. Three scenarios are indoor and the rest are outdoor.
Two of the three chosen indoor scenarios are short and challenging, while the other one is almost a straight path. The first one is a 9-meter path inside a room with a narrow space. First, the robot is controlled to follow the path after that the robot is placed at the same initial point. In the first trial, the robot follows the path with the standard funnel lane and in the second trial, it follows the path with sloped funnel lane. Figure \ref{insideroom} shows the teaching path and both paths followed by the robot with the standard and sloped funnel lane. The robot was not able to follow the path by standard funnel lane and it hits the chair. The reason is that the radius of rotation is set beforehand in the standard funnel lane and the robot turns by a constant radius. A small deviation from the desired path or switching later than it should, make it impossible to correct or compensate its direction especially in such a scenario with narrow space.

The second scenario is another 6-meter path with one turning to the left and with wide space. The robot in the repeating phase is placed two meters in front of the initial point in the teaching phase. Figure \ref{inside2} shows the followed paths with both methods. Even though in the standard funnel lane the robot constantly gets left commands, it is not able to follow the path because it is placed two meters in front of the initial point. The sloped funnel lane was able to correct its direction because it decreases its radius of rotation and it gets a sharper turning command to get back on the desired path. 

The third indoor path is almost straight 25 meters in a corridor as shown in \ref{inside3}. The results were very close and both methods followed the path successfully. 

\begin{figure}
\centering
\begin{subfigure}[b]{0.4\textwidth}
\centering 
\includegraphics[width=\linewidth]{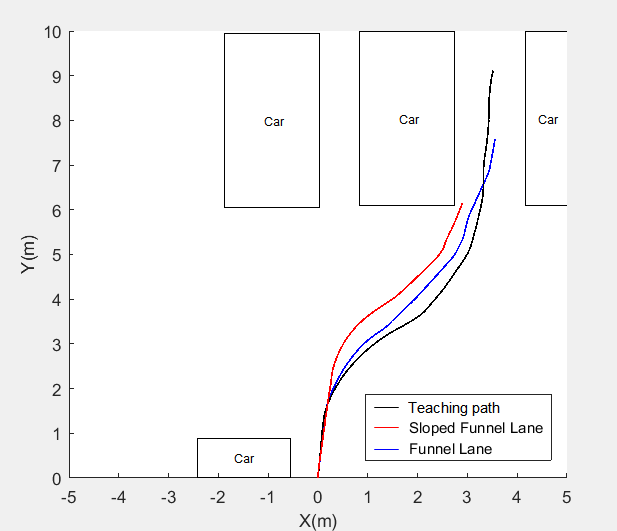}
\caption{} 
\label{parking}
\end{subfigure}\hfill
\begin{subfigure}[b]{0.4\textwidth} 
\centering
\includegraphics[width=\linewidth]{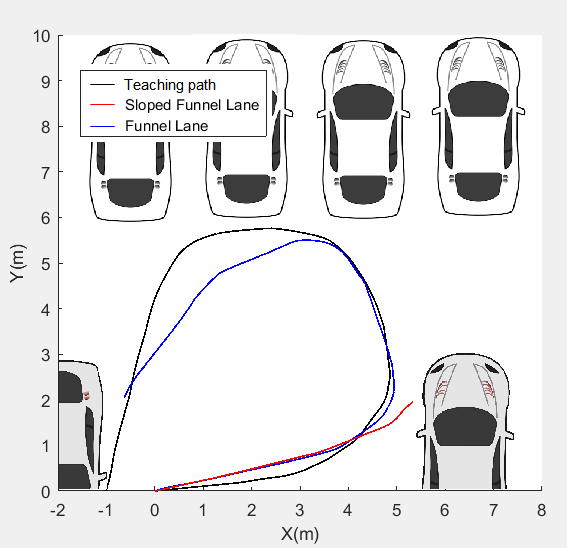}
\caption{} 
\label{loop}
\end{subfigure}\hfill
\begin{subfigure}[b]{0.4\textwidth} 
\centering
\includegraphics[width=\linewidth]{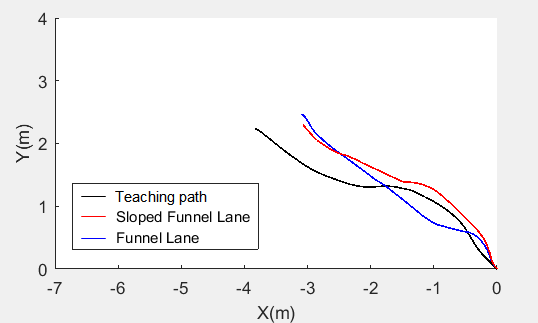}
\caption{} 
\label{wideturn}
\end{subfigure}\hfill
\caption{(a) The first outdoor parking scenario(b)The second outdoor scenario is a closed loop that two cars are left in the repeating phase (funnel lane failed) and (c) The third outdoor scenario with wide turning. }
\label{insidescenarios} 
\end{figure}

We have also chosen three outdoor scenarios. The first one is a parking lot. The robot is controlled to park between two cars near each other as shown in figure \ref{parking}. Both methods get to perform equally well. But in the standard funnel lane, the robot corrects its direction hardly and it gets closer to the side of the car which increases failure risk. Another outdoor scenario is a closed loop path with a dynamic situation. In the teaching phase, the robot is controlled to follow a looped path, and in the repeating phase two of the parked cars are left and the ability to follow the path with both methods is evaluated. Figure \ref{loop} shows the results of both methods. The gray cars are the ones left in the repeating phase. The robot failed to follow the path by the standard funnel lane because a lot of features of one side were lost and the ambiguity causes the robot to deviate and getting out the desired visual path. Last outdoor scenario is a path with wide turning and as shown in figure \ref {wideturn} both methods follow the path successfully. 

\begin{figure}
\centering
\begin{subfigure}[b]{0.2\textwidth}
\centering 
\includegraphics[width=\linewidth]{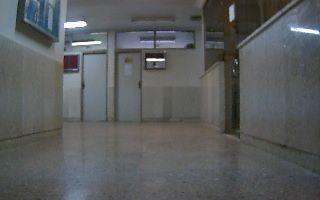}
\caption{keyframe 1} 
\label{}
\end{subfigure}\hfill
\begin{subfigure}[b]{0.2\textwidth} 
\centering
\includegraphics[width=\linewidth]{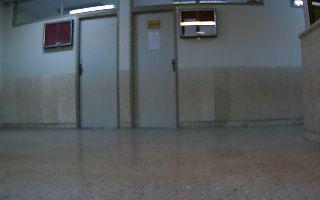}
\caption{keyframe 2} 
\label{}
\end{subfigure}\hfill
\begin{subfigure}[b]{0.2\textwidth} 
\centering
\includegraphics[width=\linewidth]{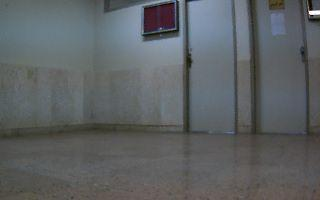}
\caption{keyframe 3} 
\label{}
\end{subfigure}\hfill
\begin{subfigure}[b]{0.2\textwidth}
\centering 
\includegraphics[width=\linewidth]{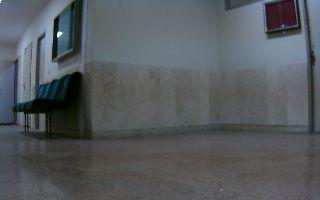}
\caption{keyframe 4} 
\label{}
\end{subfigure}\hfill
\begin{subfigure}[b]{0.2\textwidth} 
\centering
\includegraphics[width=\linewidth]{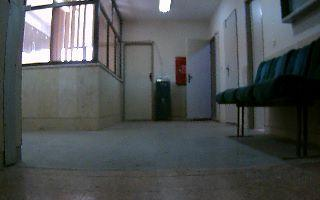}
\caption{keyframe 5} 
\label{}
\end{subfigure}\hfill
\begin{subfigure}[b]{0.2\textwidth} 
\includegraphics[width=\linewidth]{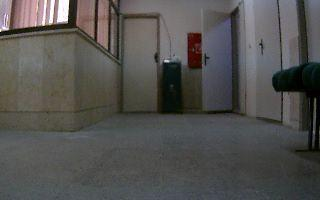}
\caption{keyframe 6} 
\label{}
\end{subfigure}\hfill
\caption{The keyframes selected to create the visual path which standard funnel lane fails to follow and sloped funnel lane follows successfully. } 
\label{inside1} 
\end{figure}
\begin{table}
\caption{The comparison of the accuracy and the repeatability of both standard funnel lane and sloped funnel lane}
\label{tab:3} 
\begin{tabularx}{0.5\textwidth}{XXX}
\hline\noalign{\smallskip}
&\textbf{standard funnel lane} &\textbf{sloped funnel lane} \\
&acc. / rep. &acc. / rep.\\
\noalign{\smallskip}\hline\noalign{\smallskip} 
\textbf{sharp turn } & 3.45 / 0.55 & \textbf{1.31 / 0.51}\\ 
\textbf{almost striaght } & 1.19 / 0.62 & \textbf{1.0 / 0.46}\\ 
\noalign{\smallskip}\hline
\end{tabularx}
\end{table}

\subsection{Accuracy and repeatability comparison}
The six practical scenarios showed the ability of both methods to follow some challenging paths. In this section, we compare the accuracy and the repeatability of both methods. The comparison method is proposed by the authors of funnel lane itself \cite{5}. Two indoor paths are chosen and the experience was repeated for ten times by both algorithms. The first one is a 10-meter path with one sharp turn to the left and low texture indoor environment. Figure \ref{inside1} shows the selected keyframes that create the visual path of the route. The second one is a 10 meter indoor almost straight route. The distance between the final point reached by the robot and the desired final point is calculated. The average RMS Euclidean distance and the standard deviation which expresses the accuracy and the repeatability of the algorithms are calculated by the following equations:
\begin{ceqn}
\begin{equation}
accuracy=\sqrt{1/n \sum_{i=1}{\parallel x_i-x_g \parallel}^2 }
\end{equation}
\begin{equation}
repeatability=\sqrt{1/n \sum_{i=1}{\parallel x_i-\mu \parallel}^2 }
\end{equation}
\end{ceqn}
where $x_g \in \Re^2$ is the desired final point and $x_i \in \Re^2$ is the final reached point and $\mu$ is:
\begin{ceqn}
\begin{equation}
\mu=1/n \sum_{i=1}{ x_i} 
\end{equation}
\end{ceqn}
The results are shown in table \ref{tab:3}. 

Actually, the robot fails to follow the path in the sharp turn with the standard funnel lane, while the sloped funnel lane was able to follow the path successfully in most cases.

It is noteworthy that the sloped funnel lane is as good as the standard funnel lane and the experiments performed shows the deficiencies of standard funnel lane has been solved successfully. 

Do not forget that in the sloped funnel lane the robot's radius of rotation is assigned adaptively, depending on the situation it faces. Therefore, the robot can deal with different turning condition including rotation in place. The robot, unlike the standard funnel lane, is free to take any path (turnings with any radius) in the teaching phase. To obviate the situations for standard funnel lane, in these experiments the robot's radius of rotation was considered almost similar and constant in both phases, however, in some cases, the standard funnel lane failed to follow them. Standard funnel lane faces a problem in turnings in narrow spacing and in sharp turnings. The reason is that the robot in such cases is facing difficulties in correcting its direction due to its constant radius of rotation. This is compounded by the impact of the ambiguity which causes the robot to deviate from the desired path.

\begin{figure}
\centering
\begin{subfigure}[b]{0.4\textwidth} 
\centering
\includegraphics[width=\linewidth]{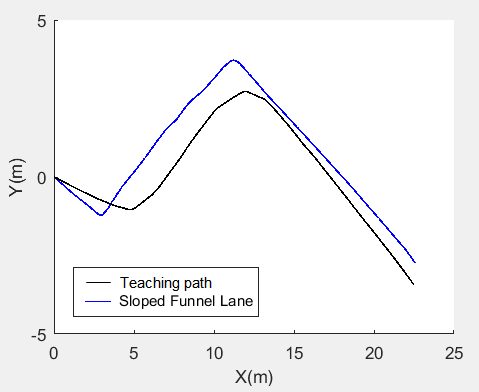}
\caption{}
\label{indoor30}
\end{subfigure}\hfill
\begin{subfigure}[b]{0.4\textwidth} 
\centering
\includegraphics[width=\linewidth]{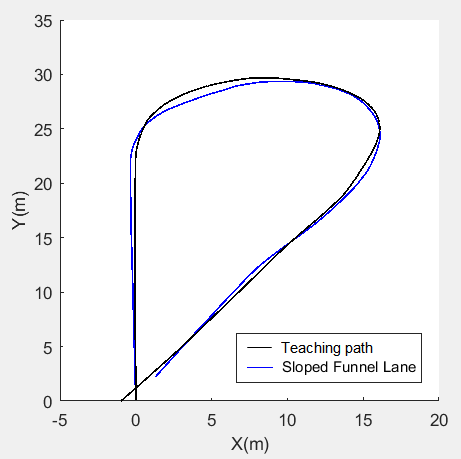}
\caption{}
\label{outdoor70}
\end{subfigure}\hfill
\caption{(a) The indoor path and (b)the outdoor path, the sloped funnel lane follow them successfully. }
\label{indoor30 and outdoor70} 
\end{figure}

Two additional experiments are conducted to demonstrate the effectiveness of the approach. The first one is a 30-meter indoor path inside the department and the second one is a 70-meter outdoor path inside IUT campus. Figure \ref{indoor30} and figure \ref{outdoor70} show the results. The most important thing in experiments is to consider the assumptions mentioned in section \ref{Notations and assumptions}. 

\section{conclusion}
\label{conclusion}
In this paper, qualitative visual navigation based on the sloped funnel lane concept was proposed. In the teaching phase, the robot is controlled manually to follow a path. In the repeating phase, the robot has to follow the desired path autonomously. First, a visual path was created by selecting some keyframes from the video taken by the robot in the teaching phase. After that in the repeating phase, the concept of the sloped funnel lane which overcomes some limitations of the standard funnel lane was introduced. The proposed sloped funnel lane, unlike the standard funnel lane, can deal with different turning conditions including rotation in place. The radius of rotation is not set beforehand which limit the maneuverability of the robot. As well it reduces the ambiguity of translation and rotation which exists in the standard funnel lane. As a result, a more robust and reliable method than the standard funnel lane has been proposed. The limitations of the standard funnel lane were explained in details and we demonstrated how the proposed sloped funnel lane overcomes them. Moreover, some experiments were conducted on a real robot and the results showed that our proposed method outperforms the standard funnel lane.

\section*{Acknowledgment}
The authors would like to thank Artificial Intelligence laboratory members for their support.

\end{document}